\documentclass[10pt, a4paper]{article}

\usepackage[final]{lrec2026} 

\usepackage{times}
\usepackage{url}
\usepackage{comment}
\usepackage{subfigure}
\PassOptionsToPackage{hyphens}{url}\usepackage{hyperref}
\usepackage{latexsym}
\usepackage{graphicx}
\usepackage{subcaption}
\usepackage[T1]{tipa}
\usepackage[english]{babel}
\usepackage{csquotes}
\usepackage{svg}

\title{Supercharging Agenda Setting Research:\\The ParlaCAP Dataset of 28 European Parliaments\\and a Scalable Multilingual LLM-Based Classification}

\name{Taja Kuzman Pungeršek$^{\ast}$, Peter Rupnik$^{\ast}$, Daniela Širinić$^{\S}$, Nikola Ljubešić$^{\ast}$$^{\dagger}$$^{\ddagger}$} 

\address{$^{\ast}$Jožef Stefan Institute; $^{\dagger}$Faculty of Computer and Information Science, University of Ljubljana;\\ $^{\ddagger}$Institute of Contemporary History; $^{\S}$Faculty of Political Science, University of Zagreb\\
          $^{\ast}$$^{\dagger}$$^{\ddagger}$Ljubljana, Slovenia; $^{\S}$Zagreb, Croatia \\
         \{taja.kuzman,peter.rupnik,nikola.ljubesic\}@ijs.si; $^{\S}$dsirinic@fpzg.hr\\}

\abstract{
This paper introduces ParlaCAP, a large-scale dataset for analyzing parliamentary agenda setting across Europe, and proposes a cost-effective method for building domain-specific policy topic classifiers. Applying the Comparative Agendas Project (CAP) schema to the multilingual ParlaMint corpus of over 8 million speeches from 28 parliaments of European countries and autonomous regions, we follow a teacher-student framework in which a high-performing large language model (LLM) annotates in-domain training data and a multilingual encoder model is fine-tuned on these annotations for scalable data annotation. We show that this approach produces a classifier tailored to the target domain. Agreement between the LLM and human annotators is comparable to inter-annotator agreement among humans, and the resulting model outperforms existing CAP classifiers trained on manually-annotated but out-of-domain data. In addition to the CAP annotations, the ParlaCAP dataset offers rich speaker and party metadata, as well as sentiment predictions coming from the ParlaSent multilingual transformer model, enabling comparative research on political attention and representation across countries. We illustrate the analytical potential of the dataset with three use cases, examining the distribution of parliamentary attention across policy topics, sentiment patterns in parliamentary speech, and gender differences in policy attention.
 \\ \newline  }

\begin{document}

\maketitleabstract

\section{Introduction}

Parliaments are at the heart of democratic governance, serving as places where elected representatives set political priorities and discuss policy issues. Understanding what parliamentarians talk about and how they allocate their attention across different policy areas provides valuable insights into how democracy works in practice. However, studying parliamentary debate agendas has been challenging, particularly when researchers want to compare multiple countries across extended time periods \citep{sebHok2023comparative}.

The Comparative Agendas Project (CAP) has been particularly influential in addressing these challenges, developing systematic ways to track political attention across different policy topics \citep{baumgartner2019comparative}. CAP provides a coherent framework for tracking media and government attention to a wide range of policy issues, enabling researchers, students, policy-makers, and journalists to examine policy-making trends over time and between countries. 
However, this research has traditionally required extensive manual work to analyze political texts, limiting the scope and scale of comparative studies.

Recent advances in computer-assisted text analysis are beginning to transform the way we study parliamentary agendas. New methods in natural language processing and machine learning, particularly transformer-based language models, make it possible to automatically analyze large collections of political texts in different languages and countries \citep{sebHok2024leveraging}. These developments offer new possibilities for understanding parliamentary debates at a much larger scale and at a much lower cost than was previously possible. However, significant challenges remain. Current datasets often cover only limited time periods and a small number of countries. 

Our project, ParlaCAP, aims to address these limitations by building on the ParlaMint corpus collection \citep{ParlaMintII-IJLRE}, the result of a CLARIN ERIC flagship project, which provides transcripts of parliamentary debates from 29 countries and autonomous regions across Europe.
\footnote{More information on the ParlaCAP project and its classifiers, datasets and tutorials is available at \url{https://clarinsi.github.io/parlacap/}.} 
We apply CAP's policy topic labels to this corpus using new multilingual natural language processing methods. Our method combines the latest large language models for annotating training datasets, which are then used for fine-tuning smaller multilingual encoder models to process millions of speeches, while maintaining the quality of human analysis, following the LLM teacher-student framework~\citep{kuzman2025llm}. The resulting ParlaCAP dataset \citep{parlacap} is developed by integrating CAP topic coding with sentiment analysis~\citep{mochtak2024parlasent} and enriching the data with metadata from PartyFacts \citep{Doring2019PartyFacts} and V-DEM \citep{vdem3} databases, with the aim to create a comprehensive resource for studying democratic politics across European parliaments. This paper shows how enhanced text-analysis approaches and techniques can transform over 8 million parliamentary speeches in more than 20 languages into structured data suitable for comparative political research. 

\section{Related Work}
\label{sec:related-work}

\citet{sebHok2023comparative} provide an overview of freely available European legislative text datasets used for large-scale comparative studies and text analysis. 
Most relevant datasets originate from a few large international projects: ParlSpeech \citep{rauh2020parlspeech}, the Comparative Agendas Project (CAP; \citealp{baumgartner2019comparative}), and ParlaMint \citep{ParlaMintI-IJLRE,ParlaMintII-IJLRE}. Among these, the ParlaMint project provides the broadest coverage of parliamentary speeches in Europe \citep{de2022language}.

Certain datasets inside the ParlaMint collection have already been enriched with policy topic labels. For example, \citet{navarrettaenriching} applied policy labels adapted from the CAP schema to the Danish ParlaMint corpus (ParlaMint-DK), part of the ParlaMint 4.1 release. The annotation combined manual labeling of agenda titles with automatic propagation of these labels to all corresponding speeches. However, this approach cannot be easily extended to all other ParlaMint corpora, as it requires manual annotation. Additionally, it is based on informative agenda titles which are not available in all ParlaMint datasets. 
This underscores the need for automated methods of topic classification using CAP labels to enable scalable annotation of speeches across all languages and corpora inside the ParlaMint collection.

A common approach to automatic topic classification involves fine-tuning deep neural transformer-based models, such as pretrained BERT-like models, on manually-annotated data. For topic classification using CAP labels, \citet{sebHok2024leveraging} introduce the CAP Babel Machine, an open-source system for classifying texts into 21 CAP policy topics across nine languages. The system is based on the XLM-RoBERTa models \citep{conneau2019unsupervised}, 
fine-tuned on manually-annotated datasets sourced from the CAP website\footnote{\url{https://www.comparativeagendas.net/datasets_codebooks}} and additional internal collections of members of the CAP network \citep{baumgartner2019comparative}. They report high performance, with weighted macro-F1 scores ranging from 0.62 to 0.96, depending on the specific domain and language. 
However, for our use case, we assume that fine-tuning a model specifically on ParlaMint data is more effective for the automatic annotation of these corpora than using existing models fine-tuned on fewer languages and different domains. We evaluate this hypothesis in Section \ref{sec:comparison-with-other-models}.

Recent advances in large language models (LLMs) have enabled alternative approaches to text classification that reduce reliance on manually-annotated data. \citet{kuzman2025llm} introduce an LLM teacher-student framework in which a state-of-the-art decoder-only LLM serves as a data annotator for news topic classification. When provided with detailed label descriptions, the LLM produces annotations comparable to those of human coders. Because LLMs are computationally expensive to deploy at scale, their role is limited to annotating training data, while smaller pretrained BERT-like student models, such as the XLM-RoBERTa model \citep{conneau2019unsupervised}, are fine-tuned on the LLM-annotated corpus. The LLM-annotated training data are shown to support strong downstream performance of the fine-tuned model. Similarly, \citet{rytting2023towards} evaluate LLMs as annotators for social science datasets. Using two English CAP datasets, they find that LLM annotations match human performance and, in some cases, exceed inter-annotator agreement among human coders. They further demonstrate the models' applicability beyond CAP topic coding, including tasks related to partisan stereotypes and populism. Building on this line of work, the present study applies the teacher-student LLM framework to develop policy topic classification models using the CAP label schema.

\begin{table*}[!ht]
\begin{center}
\begin{tabularx}{\textwidth}{|m{0.18\linewidth}|p{0.06\linewidth}|p{0.11\linewidth}|p{0.10\linewidth}|X|}
\hline
Dataset & Lang & \# Instances & \# Labels & \% Most and Least Frequent Label \\
\hline
ParlaCAP-test-EN & EN & 876 & 22 & 6.4\% (Law and Crime), 2.1\% (Culture) \\
\hline
ParlaCAP-test-HR & HR & 869 & 22 & 8.5\% (Government Operations), 1.7\% (Immigration) \\
\hline
ParlaCAP-test-SR & SR & 874 & 22 & 7.1\% (Government Operations), 1.7\% (Immigration) \\
\hline
ParlaCAP-test-BS & BS & 824 & 22 & 10.4\% (Other), 0.5\% (Culture) \\
\hline
\end{tabularx}
\caption{Information on test datasets in English (EN), Croatian (HR), Serbian (SR), and Bosnian (BS) languages, manually annotated with CAP labels.}
\label{tab:test-datasets}
    \end{center}
\end{table*}

\section{Development of the CAP Policy Topic Classifier}

In this section, we describe the development of a fine-tuned multilingual BERT-like transformer model that is specialized for the classification of CAP major labels in parliamentary speeches in ParlaMint corpora~\citep{ParlaMintII-IJLRE} by following the LLM teacher-student paradigm~\citep{kuzman2025llm}.

\subsection{CAP Labels}

To follow the tradition of the Comparative Agendas Project, we use the 21 major CAP topics as they are defined in the CAP Master Codebook~\citep{bevan2019gone}.\footnote{\url{https://www.comparativeagendas.net/pages/master-codebook}} An inspection of the ParlaMint corpora revealed that some speeches do not address a policy topic, for example, when speakers discuss meeting logistics, share personal stories, or engage in arguments. To account for these cases, we introduced an additional label, \textit{Other}. Thus, the final label set comprises 22 categories. The label descriptions provided to the large language model and annotators (see Section \ref{subsec:appendix-labels} in appendix) were developed from the list of subtopics and their descriptions in the Master Codebook, and were further extended using the Croatian CAP guidelines \citep{sirinic_2019} and expert input.

\subsection{LLM-Annotated Training Data}
\label{sec:training-data}

To specialize the model for classification of parliamentary speeches in ParlaMint corpora \citep{ParlaMintI-IJLRE}, the training dataset is constructed from all corpora in the ParlaMint 4.1 corpus collection \citep{parlamint-4.1}. ParlaMint 4.1 is a collection of comparable corpora that contain transcriptions of parliamentary debates of 29 European countries and autonomous regions, mostly covering the period from 2015 to mid-2022. The individual corpora comprise between 9 and 126 million words, and the complete corpus collection contains more than 1.2 billion words. 

The development of the training data consists of two steps: 1) preparation of suitable samples of speeches from each ParlaMint dataset, and 2) automatic annotation of the training dataset with a large language model. More specifically, we sample 1,200 instances per each of 29 ParlaMint corpora, resulting in a dataset of 34,800 speeches. Annotation with CAP labels is performed automatically using the GPT-4o model \citep{openai-gpt4o}, following the prompt and label description that are provided in the appendix (see Sections \ref{subsec:appendix-labels} and \ref{subsec:appendix-prompt}). Prediction through the OpenAI API batch option cost approximately \$100. The suitability of using an LLM as a data annotator is evaluated in Section \ref{sec:test-data}.



An analysis of the resulting training data revealed a severe underrepresentation of the \textit{Public Lands} category, which appeared only in around hundred training instances. To address this, we implement a targeted data augmentation pipeline inspired by the LLM-based ``finding a needle in a haystack'' approach \citep{mochtak2026needle}. We define a list of keywords associated with the \textit{Public Lands} category based on its definition (e.g., ``national park'', ``forest fire'', ``grazing''). Since the ParlaMint corpora have been machine-translated into English, we can identify speeches containing specific keywords across all ParlaMint languages by searching for them in the English ParlaMint.en-ana \citep{parlamint-en-ana} corpus available through the \href{https://www.clarin.si/ske/#dashboard?corpname=parlamint41_xx_en}{CLARIN.SI concordancer}. With this approach, we extract up to 2,000 random candidate instances per keyword across all ParlaMint corpora. 
Using the GPT-4o model with the same prompt as for the annotation of the training data, we annotate the candidate speeches and identify 779 as belonging to the \textit{Public Lands} category. These are added to the training data, increasing the total number of examples for this label from 145 to 924.

The final training dataset \citep{parlacap-train} consists of 35,579 speeches. It includes a training split (29,779 instances -- 1,000 per corpus and the additional 779 \textit{Public Lands} instances) and a development split (5,800 instances -- 200 per corpus), with stratification performed before augmenting the \textit{Public Lands} label. The dataset is freely accessible in the CLARIN.SI repository\footnote{The training dataset can be downloaded from the CLARIN.SI repository: \url{http://hdl.handle.net/11356/2093}.}.

\begin{figure}[t]
  \centering
  \includegraphics[width=1\columnwidth]{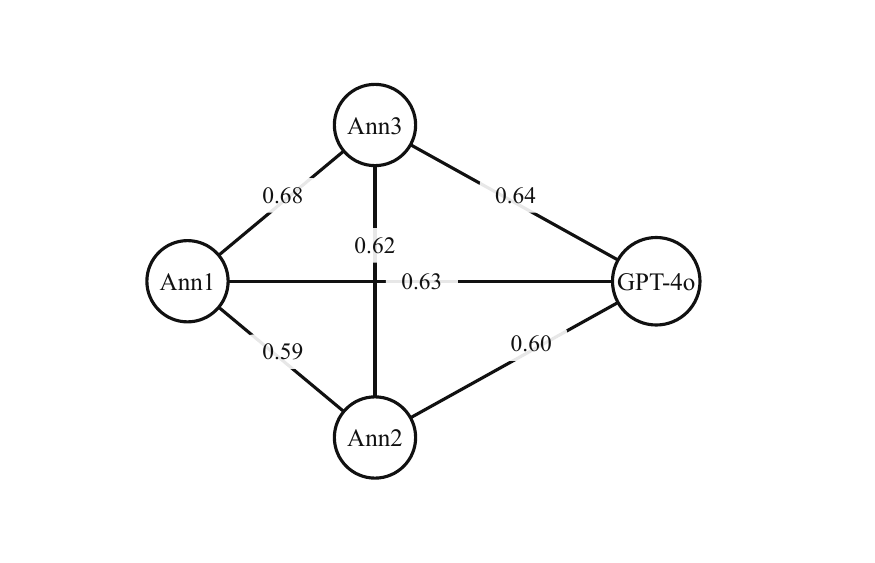}
  \caption{The inter-annotator agreement in terms of nominal Krippendorff's alpha between human annotators and the large language model (GPT-4o).}
  \label{fig:inter-annotator-agreement}
\end{figure}

\subsection{Manually-Annotated Test Data}
\label{sec:test-data}

We construct test datasets in four languages, namely English, Croatian, Serbian, and Bosnian. The test instances are extracted from the respective ParlaMint 4.1 corpora \citep{parlamint-4.1}, that is, ParlaMint-GB 4.1, ParlaMint-HR 4.1, ParlaMint-RS 4.1 and ParlaMint-BA 4.1. 
From each corpus, we randomly sample 10,000 speeches, ensuring that none of them overlaps with the training dataset. Then we automatically annotate the sample with the GPT-4o model using the CAP labels following the same methodology as for annotating the training dataset. To create a label-balanced test dataset, we sample 40 speeches per CAP label from the 10,000 speeches automatically annotated by the LLM, resulting in test datasets of approximately 800--880 instances per language.

The test datasets are manually annotated by an expert annotator. This annotator was selected from a group of three initial annotators who annotated a sample of the Croatian test dataset. The annotator with the highest inter-annotator agreement with the other two was chosen to annotate the remaining test datasets. The annotators had prior experience with news topic or policy topic annotation. They followed the annotation guidelines and label descriptions (see Sections \ref{subsec:appendix-labels} and \ref{subsec:appendix-guidelines} in the appendix) and completed brief training. In addition to the 21 CAP main labels and the label \textit{Other}, annotators could assign the label \textit{do not know} to texts with unclear topics. These instances were excluded from the final test datasets. Table \ref{tab:test-datasets} provides details on the size of each test dataset and the distribution of manually-annotated labels within them. The datasets are approximately balanced across labels.

To evaluate the annotator performance and compare it with the performance achieved by the LLM on this task, about 400 instances of the Croatian test dataset were manually annotated by three human annotators. The inter-annotator agreement is calculated using the Krippendorff’s alpha metric \citep{krippendorff2018content}. The evaluation shows that the agreement between the three annotators ranges from 0.59 to 0.68, while the agreement between annotators and the GPT-4o model ranges from 0.60 to 0.64, as shown in Figure \ref{fig:inter-annotator-agreement}. The results confirm that the LLM performs as reliably as human annotators on this task, which is consistent with findings from similar topic classification tasks \citep{kuzman2025llm}, and supports its use for annotating the training data.

In addition to serving for the evaluation of the fine-tuned CAP classification models (see Section \ref{sec:evaluation}), the test datasets can also be used to assess the performance of large language models on this task. They have already been used in a study that compared fine-tuned BERT-like models with various closed-source and open-source large language models in a zero-shot setting \citep{pungervsek2025state}. Accordingly, to preserve the integrity of future evaluations, the datasets are not publicly released, as this would risk their inclusion in LLM pretraining or fine-tuning. Access to the test datasets can be granted upon request from the corresponding authors.

\subsection{Model Fine-Tuning}

To identify the best performing model for our task, we experiment with fine-tuning two multilingual pretrained BERT-like models, namely, the large-size XLM-RoBERTa model\footnote{\url{https://huggingface.co/xlm-roberta-large}} \citep{conneau2019unsupervised} and the XLM-R-Parla model\footnote{\url{https://huggingface.co/classla/xlm-r-parla}} \citep{xlm-r-parla}. The XLM-R-Parla model is based on the XLM-RoBERTa-large architecture \citep{conneau2019unsupervised} and was further pretrained on 1.72 billion words from parliamentary proceedings in 30 European languages, using data from the ParlaMint 3.0 \citep{Parlamint30} and EuroParl \citep{europarl} corpora. Previous experiments on sentiment identification showed that additional pretraining on parliamentary domain data improves performance \citep{mochtak2024parlasent}.

Both models are fine-tuned on the initial ParlaMint training split (29,000 instances) that was created before the augmentation with the \textit{Public Lands} instances. The optimal hyperparameters were identified via a hyperparameter search performed on the development dataset. We use a learning rate of $1 \times 10^{-5}$ and 3 epochs, and save each model variant three times to enable significance testing.

Initial experiments reveal low performance on the \textit{Public Lands} category due to data sparsity. We thus perform another experiment where we fine-tune the XLM-R-Parla model on the training dataset to which we have added additional instances of the \textit{Public Lands} (see Section \ref{sec:training-data}), totaling 29,779 instances. We observe a substantial improvement of the model's performance on the \textit{Public Lands} label, with its F1 score on this label increasing from 0.30 to 0.80. 


\subsection{Evaluation Results}
\label{sec:evaluation}

\begin{table*}[!ht]
\begin{center}
\begin{tabularx}{\textwidth}{|l|l|X|X|X|X|}
      \hline
Model       & Training Data     & EN      & HR     & SR      & BS      \\
\hline
\hline
GPT-4o & - & 0.74 & 0.68 & 0.72 &	0.63 \\
\hline
\hline
XLM-R-Parla & ParlaMint+PL & 0.72$\pm$0.01 & 0.68$\pm$0.01 & 0.71$\pm$0.01 & 0.64$\pm$0.01 \\
      \hline
XLM-R-Parla & ParlaMint     & 0.70$\pm$0.01 & 0.66$\pm$0.01 & 0.68$\pm$0.01 & 0.62$\pm$0.01 \\
      \hline
XLM-RoBERTa & ParlaMint     & 0.68$\pm$0.01 & 0.65$\pm$0.01 & 0.67$\pm$0.01 & 0.63$\pm$0.01 \\
      \hline
\end{tabularx}
    \caption{Performance of fine-tuned models on English (EN), Croatian (HR), Serbian (SR) and Bosnian (BS) test data in terms of macro-F1. Each model was fine-tuned and evaluated three times. All models were fine-tuned on the training data from the ParlaMint datasets (see Section \ref{sec:training-data}). In one setting (ParlaMint+PL), the training data were extended with additional instances of the \textit{Public Lands} label. Performance of the GPT-4o model is added as an upper threshold as the models were fine-tuned on the training data that was annotated by the GPT-4o model.}
    \label{tab:results-evaluation}
\end{center}
\end{table*}

The results presented in Table \ref{tab:results-evaluation} demonstrate the effectiveness of domain-specific pretraining and targeted data augmentation for classification tasks in the parliamentary domain.

The XLM-R-Parla model, which is based on the XLM-RoBERTa-large architecture and further pretrained on multilingual parliamentary corpora, slightly outperforms the baseline XLM-RoBERTa model on three out of four test datasets.

Further improvements are observed with XLM-R-Parla fine-tuned on extended training data that incorporate additional LLM-annotated \textit{Public Lands} instances. It achieves macro-F1 scores above 0.70 on English and Serbian, and scores between 0.64 and 0.68 on Bosnian and Croatian test datasets, respectively. This model outperforms the original fine-tuned XLM-R-Parla and XLM-RoBERTa models on all test datasets.

As shown in Table \ref{tab:results-evaluation}, the models perform the worst on the Bosnian test dataset. A manual analysis of the dataset, together with feedback from the annotator, indicates that this dataset is more challenging due to less structured debates and a higher frequency of vague borderline cases in the Bosnian parliament. Notably, the most frequent label in this dataset is \textit{Other}, which covers topics unrelated to policy agendas, such as personal stories, interjections, and exchanges between members. Similar to that, the final Bosnian parliament annotations have by far the largest proportion of speeches annotated with the \textit{Government Operations} class, which is another signal for the specificity of the Bosnian dataset.

The best-performing fine-tuned XLM-R-Parla model achieves performance comparable to the much larger GPT-4o model, while being more scalable and computationally efficient. This highlights the effectiveness of the LLM teacher-student approach that enabled the development of a scalable domain-specific classifier by leveraging a larger LLM as a data annotator. Moreover, the fine-tuned model achieves a performance comparable to that of human annotators. As shown in Figure \ref{fig:inter-annotator-agreement}, the level of agreement between the human annotators is similar to their agreement with the LLM model, and the best-performing student BERT-like model attains performance comparable to that of the LLM.

\subsection{ParlaCAP Model}
\label{sec:parlacap-classifier}

We publish the best-performing fine-tuned model in the Hugging Face repository under the name ParlaCAP classifier \citep{parlacap_model}.\footnote{The ParlaCAP classifier is available in Hugging Face at \url{https://www.doi.org/10.57967/hf/6684}.} The published model is based on the multilingual parliamentary XLM-R-Parla BERT-like model that was fine-tuned on LLM-annotated ParlaMint training data in 29 languages, extended with additional \textit{Public Lands} instances (29,779 speeches).

For downstream applications,
\footnote{The code for applying the ParlaCAP classifier to ParlaMint-style datasets is freely available on GitHub at \url{https://github.com/clarinsi/ParlaMint-Annotation-with-CAP-Topics}.} 
we recommend incorporating a confidence-based filtering mechanism to further improve the reliability of model predictions. Specifically, to annotate the entire ParlaMint 5.0 corpus collection with the ParlaCAP classifier, we apply a confidence threshold of 0.60 and label any instance below this threshold as \textit{Mix}, indicating that the model is uncertain about the dominant policy topic.

\begin{table}[!ht]
\begin{center}
\begin{tabularx}{\columnwidth}{|l|X|X|X|}
      \hline
Dataset   & Micro-F1 & Macro-F1 & Accuracy  \\
      \hline
EN & 0.76    & 0.76    & 0.76              \\
      \hline
HR & 0.72    & 0.73    & 0.72                \\
      \hline
SR & 0.75    & 0.74    & 0.75                \\
      \hline
BS & 0.69    & 0.68    & 0.69               \\
      \hline
\end{tabularx}
    \caption{Performance of the final published ParlaCAP policy topic classifier on English (EN), Croatian (HR), Serbian (SR) and Bosnian (BS) test data in terms of micro-F1, macro-F1 and accuracy. The evaluation includes only instances with a prediction confidence score above 0.60, which account for 90\% of all test instances. }
    \label{tab:parlacap-final-results}
 \end{center}
\end{table}

When evaluated on the test datasets, this strategy results in 9\% of the instances being marked as \textit{Mix} in the English test set and 11\% in the other three test sets. Importantly, by excluding these low-confidence predictions, the model achieves strong performance on the remaining data, ranging from 0.69 of micro-F1, macro-F1 and accuracy in Bosnian to 0.76 in English (see Table \ref{tab:parlacap-final-results}).

\begin{figure}[!ht]
\begin{center}
        \includegraphics[width=\columnwidth]{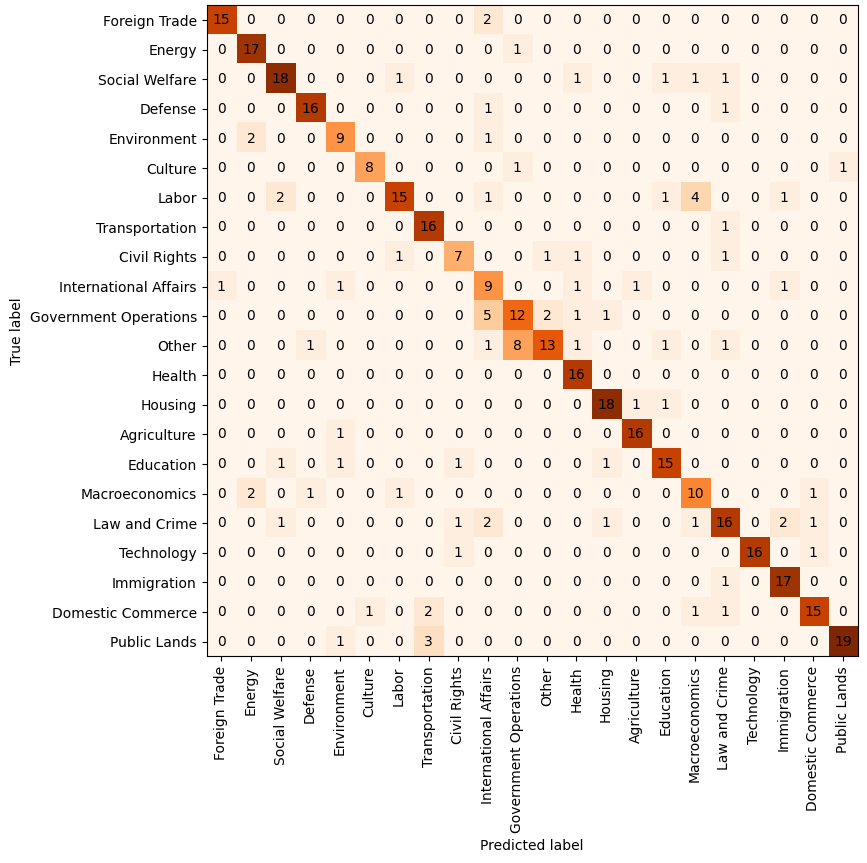}
        \caption{Performance of the ParlaCAP model on the English test dataset after removal of \textit{Mix} instances predicted with low confidence.}
        \label{fig:cm-en}
\end{center}
\end{figure}

The published model achieves high performance on all labels, as shown in Figure \ref{fig:cm-en} that presents the performance on the English test dataset.

\subsection{Comparison to Other CAP Models}
\label{sec:comparison-with-other-models}

We finally evaluate other publicly available models fine-tuned to the CAP schema by \citet{sebHok2024leveraging}, presented in Section \ref{sec:related-work}.\footnote{The models are available in Hugging Face at \url{https://huggingface.co/collections/poltextlab/cap-babel-672b46e9d40b55aa6f55da3e}.} These models are based on the large-sized XLM-RoBERTa pretrained model \citep{conneau2019unsupervised} and have been fine-tuned on different domains of manually-annotated CAP datasets in various languages. 
The models use the same CAP major labels as our models, and the category \textit{No Policy Content} which has been mapped to our category \textit{Other}.

\begin{table*}[!ht]
\begin{center}
\begin{tabularx}{\textwidth}{|l|X|X|X|X|}
      \hline
Model                                       & EN & HR & SR & BS \\ \hline
ParlaCAP & 0.72    & 0.69     & 0.71    & 0.65    \\ \hline
\href{https://huggingface.co/poltextlab/xlm-roberta-large-party-cap-v3}{xlm-roberta-large-party-cap-v3}              & 0.64    & 0.63     & 0.62    & 0.57    \\ \hline
\href{https://huggingface.co/poltextlab/xlm-roberta-large-parlspeech-cap-v3}{xlm-roberta-large-parlspeech-cap-v3}         & 0.64    & 0.57     & 0.60    & 0.55    \\ \hline
\href{https://huggingface.co/poltextlab/xlm-roberta-large-pooled-cap-v3}{xlm-roberta-large-pooled-cap-v3}             & 0.64    & 0.60     & 0.62    & 0.57    \\ \hline
\href{https://huggingface.co/poltextlab/xlm-roberta-large-execspeech-cap-v3}{xlm-roberta-large-execspeech-cap-v3}         & 0.60    & 0.58     & 0.60    & 0.53    \\ \hline
\href{https://huggingface.co/poltextlab/xlm-roberta-large-execorder-cap-v3}{xlm-roberta-large-execorder-cap-v3}          & 0.60    & 0.59     & 0.58    & 0.51    \\ \hline
\href{https://huggingface.co/poltextlab/xlm-roberta-large-legislative-cap-v3}{xlm-roberta-large-legislative-cap-v3}        & 0.59    & 0.62     & 0.61    & 0.56    \\ \hline
\href{https://huggingface.co/poltextlab/xlm-roberta-large-media-cap-v3}{xlm-roberta-large-media-cap-v3}              & 0.57    & 0.56     & 0.58    & 0.49    \\ \hline
\href{https://huggingface.co/poltextlab/xlm-roberta-large-social-cap-v3}{xlm-roberta-large-social-cap-v3}             & 0.57    & 0.53     & 0.51    & 0.44    \\ \hline
\href{https://huggingface.co/poltextlab/xlm-roberta-large-budget-cap-v3}{xlm-roberta-large-budget-cap-v3}             & 0.52    & 0.51     & 0.48    & 0.46    \\ \hline
\end{tabularx}
        \caption{Performance of the ParlaCAP model, in comparison to the performance of other existing fine-tuned policy topic models by \citet{sebHok2024leveraging}. The models are evaluated on English (EN), Croatian (HR), Serbian (SR) and Bosnian (BS) test data in terms of macro-F1. No confidence-based filtering is applied, i.e., ParlaCAP predictions are evaluated as they were produced.}
    \label{tab:comparison-to-cap-models}
 \end{center}
\end{table*}

Since both the ParlaCAP model and other evaluated models are based on similar pretrained models, the performance differences shown in Table \ref{tab:comparison-to-cap-models} mainly reflect the similarity between their fine-tuning data and our test datasets. For our specific objective -- enriching the ParlaMint datasets with topic information -- we demonstrate that fine-tuning on data closely aligned with the target domain is advantageous: the ParlaCAP model outperforms models fine-tuned on data from other domains.


\section{The ParlaCAP Dataset}

The ParlaCAP classifier, introduced in Section \ref{sec:parlacap-classifier}, is applied to the ParlaMint 5.0 corpus collection \citep{parlamint5.0}, resulting in the ParlaCAP dataset \citep{parlacap}. This dataset builds on the speeches and metadata provided in the ParlaMint corpora, and adds topic and sentiment labels, as well as additional metadata. In contrast to the ParlaMint corpora, which are distributed in formats tailored to linguistic analyses, the ParlaCAP dataset is provided in a tabular format that follows the ``text-as-data'' paradigm and is designed to meet the needs of social scientists.

\begin{figure}[!ht]
\begin{center}
        \includegraphics[width=\columnwidth]{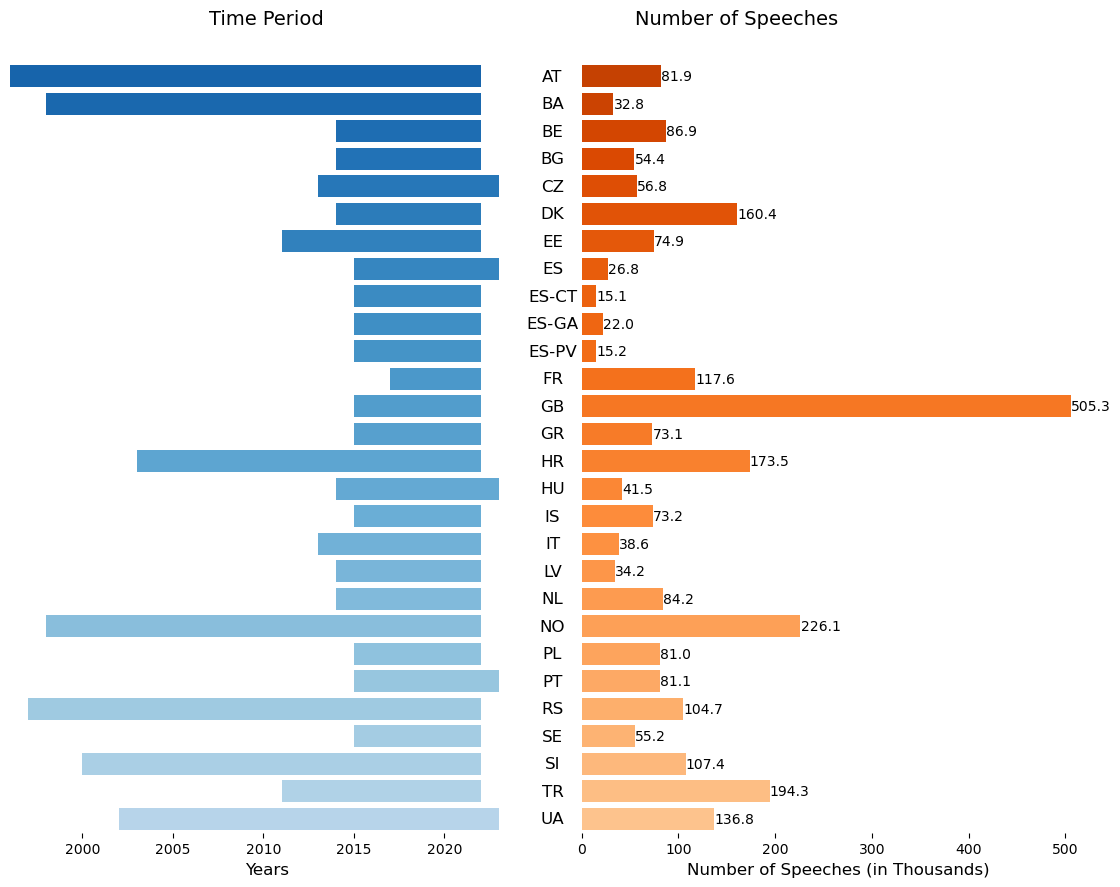}
        \caption{Scope of the ParlaCAP dataset, showing the number of speeches in each parliamentary dataset and their temporal coverage.}
        \label{fig:dataset}
\end{center}
\end{figure}

The ParlaCAP dataset comprises 8 million speeches from 28 European national and regional parliaments (see Figure \ref{fig:dataset}), namely, Austrian (AT), Bosnian (BA), Belgian (BE), Bulgarian (BG), Czech (CZ), Danish (DK), Estonian (EE), Spanish (ES), Catalan (ES-CT), Galician (ES-GA), Basque (ES-PV), French (FR), British (GB), Greek (GR), Croatian (HR), Hungarian (HU), Icelandic (IS), Italian (IT), Latvian (LV), Dutch (NL), Norwegian (NO), Polish (PL), Portuguese (PT), Serbian (RS), Swedish (SE), Slovenian (SI), Turkish (TR), and Ukrainian (UA). Each speech is assigned a sentiment label -- negative, neutral, or positive -- using the ParlaSent classifier for sentiment analysis in parliamentary texts (\citealp{parlasent-model}, \citealp{mochtak2024parlasent}). In addition, each speech is labeled with a policy topic according to the CAP schema, using the ParlaCAP classifier presented in this paper. The dataset also includes extensive metadata on speakers, political parties, and democratic contexts.

The dataset is freely available in the CROSSDA repository under the CC BY 4.0 license.
\footnote{The ParlaCAP dataset is available at \url{https://doi.org/10.23669/1ZTELP}.} 
For each parliament, it is distributed in three tabular (TSV) files. The first file is a speech-level TSV that contains the full speech text (along with the text machine-translated to English), the assigned topic label, and the aggregated sentiment label. It also includes rich metadata from ParlaMint (e.g., speaker, party, party status, chairing role), as well as additional identifiers such as the PartyFacts Party ID \citep{Doring2019PartyFacts} and the V-Dem Country ID \citep{vdem3}. The second file is a speech-level TSV without speech text, which reduces the file size by approximately 88\% and facilitates more efficient quantitative analyses. The third file is a sentence-level TSV that provides the speech ID, the sentiment label assigned to each sentence, and the sentence text. This structure allows users to conduct analyses at both the speech and sentence levels, depending on their research needs.\footnote{Tutorials for analyzing parliamentary speeches from multiple European countries using Python and R, based on the ParlaCAP dataset, are available at \url{https://github.com/clarinsi/ParlaCAP-Analysis-Tutorials}.}

\section{Use Cases} 

The ParlaCAP dataset opens new possibilities for studying fundamental questions in democratic governance and political representation. We illustrate the analytical potential of the ParlaCAP dataset with three examples: examining the distribution of parliamentary attention across policy topics, 
exploring sentiment patterns in parliamentary speech, and showing gender differences in policy attention.

For temporal consistency across parliaments, we include here data from years 2017--2022. We include only speeches given by members of parliament and discard speeches of chairpersons. Speeches, annotated with labels \textit{Other} and \textit{Mix}, are omitted from the analysis.


\subsection{What Do Parliamentarians Talk About?}

\begin{figure}[!ht]
\begin{center}
\includegraphics[width=\columnwidth]{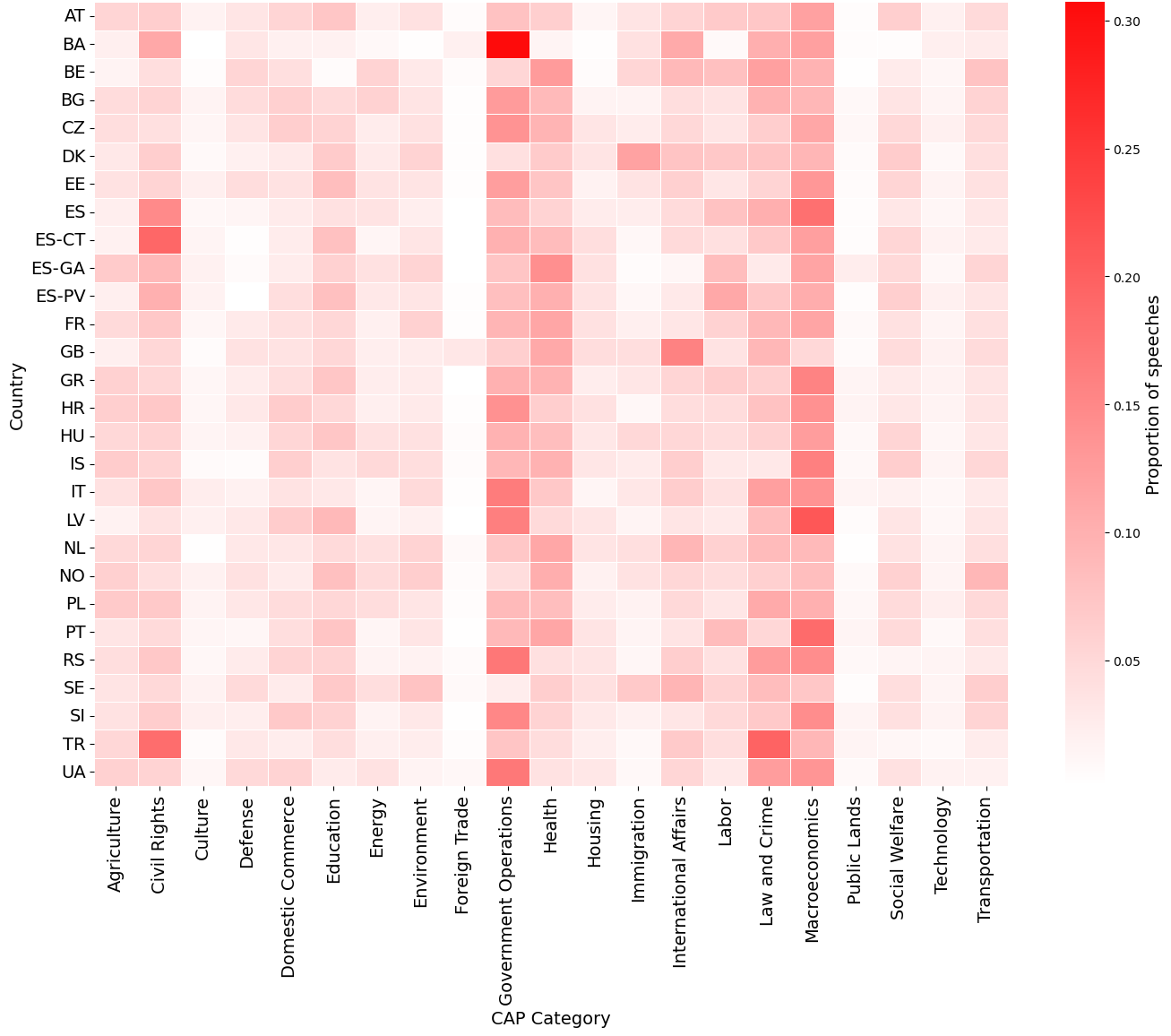}
        \caption{Probability distribution of automatically annotated CAP labels for each parliament included in the ParlaCAP dataset.}
        \label{fig:parlacap-analysis-label-distribution}
\end{center}
\end{figure}

Figure \ref{fig:parlacap-analysis-label-distribution} presents an illustrative analysis of the ParlaCAP dataset, demonstrating the research potential of applying the CAP topic classification to parliamentary speeches across Europe. The visualization shows the distribution of parliamentary attention across 21 policy topics, with values representing the percentage of total parliamentary speeches devoted to each area (ranging from 0.0 to 0.25, or 0\% to 25\%).

Certain policy areas, \textit{Macroeconomics}, \textit{Government Operations}, and \textit{Health}, are consistently appearing as high-attention topics across most parliaments, reflecting shared challenges. Other areas, such as \textit{Culture}, \textit{Foreign Trade}, and \textit{Public Lands} receive relatively limited focus across countries. However, attention intensity and distribution varies significantly across countries, with a parliamentary focus ranging from highly concentrated (darker red patterns in Figure \ref{fig:parlacap-analysis-label-distribution}) to more dispersed engagement across topics. This variation demonstrates that national political contexts, institutional structures, and domestic priorities shape the parliamentary agenda. 

These patterns only point to deeper dynamics of agenda competition and prioritization. When core functions of government -- such as \textit{Macroeconomics} and \textit{Government Operations} -- dominate parliamentary debates, legislatures appear to pursue a less diverse agenda, focusing the majority of their attention on fewer critical issues. This echoes broader findings from \citet{Jennings2011} about executive agenda setting, where some issues prove to be ``more equal than others'' in the competition for political attention. Topics like \textit{Culture}, \textit{Public Lands}, and \textit{Foreign Trade} might be secondary concerns that receive focus only when urgent governance pressures ease. This suggests hierarchical attention allocation. Some issues are permanent fixtures of parliamentary debate, while others are allowed to appear on the floor only when political space allows. However, more rigorous empirical tests are required to validate these descriptive observations. 

\subsection{How Do Parliamentarians Talk?}

\begin{figure}[!ht]
\begin{center}
\includegraphics[width=\columnwidth]{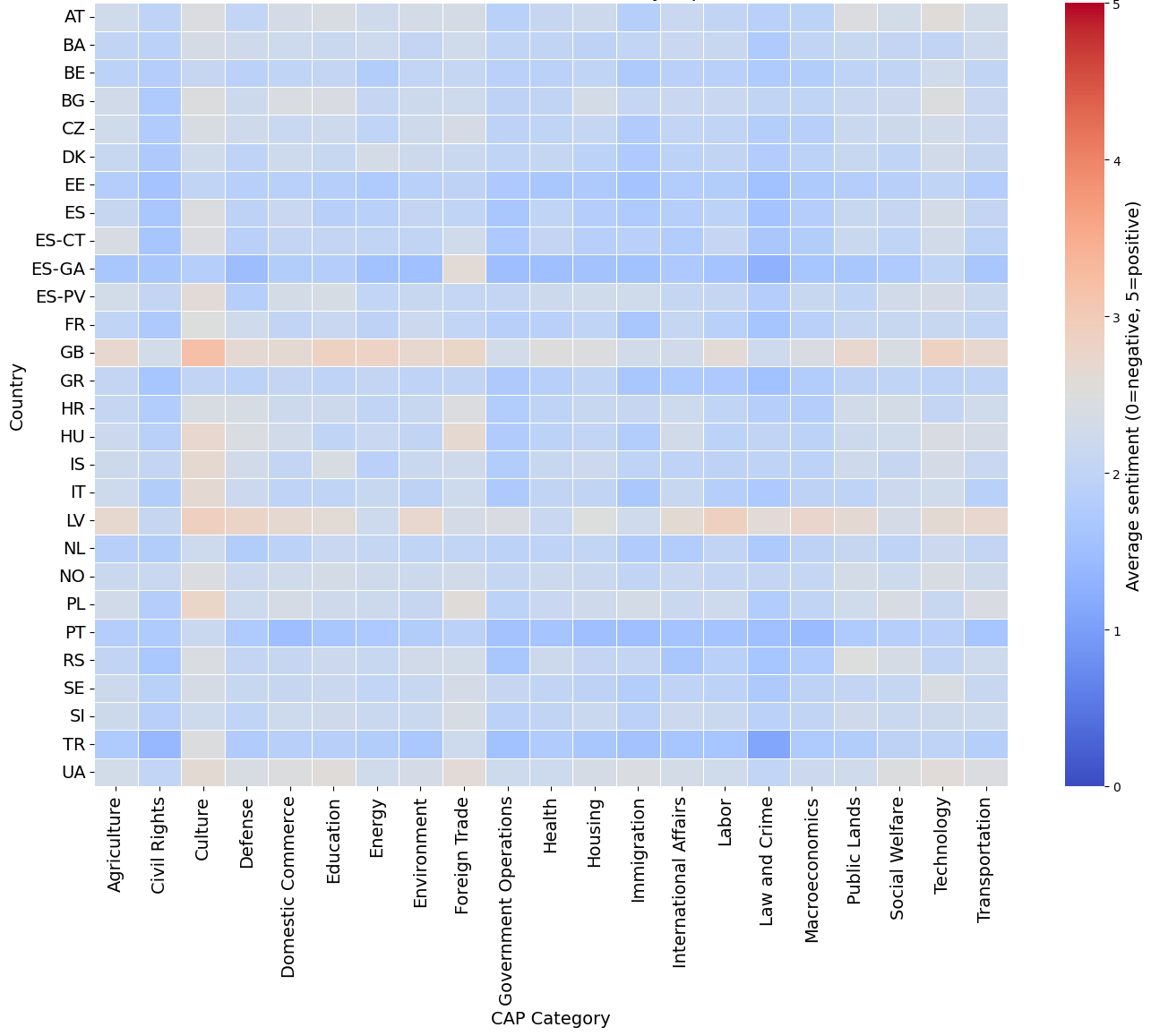} 
        \caption{Mean sentiment across CAP topics, sentiment ranging from 0 (negative) to 5 (positive). Red denotes more positive sentiment and blue representing more negative sentiment towards the topic.}
        \label{fig:sentiment-across-topics}
\end{center}
\end{figure}

The ParlaCAP dataset allows us to study not just what politicians talk about, but how they talk about it. It has been enhanced with sentiment analysis, revealing the emotional tone of parliamentary debates across different policy topics in 28 European countries. 
In Figure \ref{fig:sentiment-across-topics} the colors represent sentiment scores from 1.2 to 3.2, where darker blue indicates more negative language and lighter colors show more positive tone. 

The analysis reveals several clear patterns. Most European parliaments use predominantly negative language across policy areas, as shown by the widespread blue coloring. This finding connects to recent research by \citet{poljak2024}, who showed that politicians frequently use negative rhetoric because negative language attracts media attention. According to his theory, negative stories engage audiences more effectively, creating a cycle where politicians use more negativity to get more media coverage.

If we look at separate topics in Figure \ref{fig:sentiment-across-topics} we see several distinct patterns. \textit{Culture} receives the most positive treatment across countries, likely because cultural discussions tend to be celebratory and unifying. \textit{Law and Crime} topics are consistently the most negative, which makes sense given these debates often emerge as a reaction on current problems, such as criminal behavior or punishment debates.

Great Britain is noticeably different, showing more positive sentiment across most topics compared to other European countries. This might reflect unique aspects of British parliamentary culture or different media dynamics that do not reward negativity as strongly.

\subsection{Do Men and Women Talk about Different Issues?}

\begin{figure}[!ht]
\begin{center}
\includegraphics[width=\columnwidth]{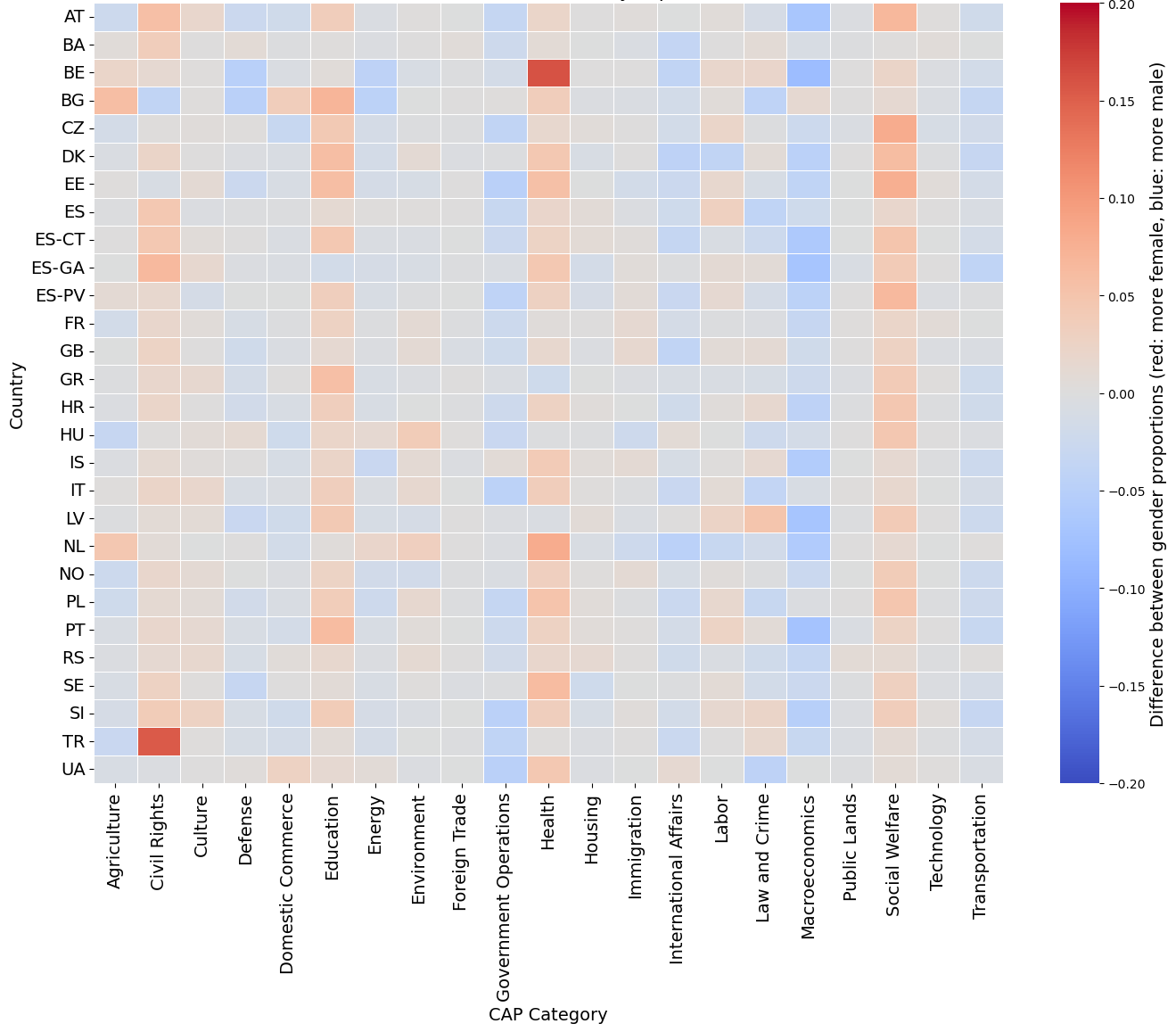}
        \caption{The absolute difference in topic probability between female and male parliament-level distributions. Red color signals that female PMs discuss the specific topic in the specific parliament more, while blue color signals the opposite.}
        \label{fig:gender-across-topics}
\end{center}
\end{figure}

Figure \ref{fig:gender-across-topics} demonstrates another dimension of the analytical richness of the ParlaCAP dataset by incorporating speaker gender metadata from ParlaMint. The visualization shows the difference in topic attention between female and male parliamentarians, with red indicating topics where women speak proportionally more and blue showing areas where men dominate the debate. Gender metadata originally take three values: "M", "F", and "U" (unknown), with only 10 speeches from Spain being coded with "U". 

The results in Figure \ref{fig:gender-across-topics} align with theoretical expectations about gendered political priorities and role specialization in parliamentary work \citep{deVet2023}. Women speak more often about \textit{Health}, \textit{Social Welfare}, and \textit{Education} topics across most countries, as shown by the red coloring in these areas. Men have a stronger presence in areas like transportation, economics, and defense-related topics, which aligns with traditional associations between masculinity and ``hard'' policy domains involving infrastructure, finance, and security.

In some countries, we see larger gender gaps than in others, showing significant and interesting cross-national variation (notable outliers include Turkey with particularly strong gender differences in \textit{Civil Rights}).

\section{Conclusion}

In this paper, we make two contributions: we demonstrate an effective and computationally practical method for domain-specific topic classification; and we introduce ParlaCAP, a dataset providing speech-level policy topic information for 28 European parliaments.

The ParlaCAP classifier 
 \citep{parlacap_model}\footnote{The ParlaCAP topic classifier is available in Hugging Face at \url{https://www.doi.org/10.57967/hf/6684}.}
is a domain-specific multilingual transformer model for topic classification in parliamentary texts based on the Comparative Agendas Project (CAP) schema. The classifier is developed within the LLM teacher-student framework \citep{kuzman2025llm}, which enables scalable fine-tuning of BERT-like models without relying on manually-annotated training data. 

We show that large language models can replace manual annotation for building domain-specific topic classifiers. Our approach is simple: use a large language model to annotate a subset of parliamentary speeches, then fine-tune a smaller BERT-like model on this LLM-annotated training data. This procedure produces a classifier that is specialized for the specific dataset of interest. We show that the LLM can serve as a reliable annotator for policy topic classification, as its agreement with human annotators is comparable to inter-annotator agreement among humans. Moreover, smaller models fine-tuned on LLM-annotated in-domain data achieve performance comparable to the LLM itself. Compared to existing models that were fine-tuned on manually-annotated data from different datasets and domains, our approach performs better on the target parliamentary data. This finding highlights the importance of domain-aligned training data: a model trained on in-domain data annotated by an LLM can outperform models trained on manually-annotated but out-of-domain data. At the same time, the approach remains cost-effective, as manual annotation is required only for evaluation rather than for large-scale training.

Secondly, this paper introduces the ParlaCAP dataset \citep{parlacap},\footnote{The ParlaCAP dataset is available in the CROSSDA repository at \url{https://doi.org/10.23669/1ZTELP}.} which represents a major step forward in comparative political research, providing unprecedented access to systematic analysis of parliamentary attention in 28 countries and autonomous regions. With speech-level data on parties, speakers, topics, and sentence-level data on sentiment, this freely-available dataset provides new opportunities for political scientists and social science researchers to explore how parliaments work. 

\section{Acknowledgments}

We would like to thank the annotators of the test datasets, especially Mirna Potočnjak, the main annotator, for their diligence and the time devoted to manual annotation, which resulted in the high-quality evaluation datasets used in this work. We would also like to thank the \href{https://www.clarin.si/info/k-centre/}{CLASSLA knowledge centre for South Slavic languages} and the Slovenian \href{https://www.clarin.si/info/about/}{CLARIN.SI infrastructure} for their valuable support.

This work was supported in part by the project ``Large Language Models for Digital Humanities'' (Grant GC-0002), the research programme ``Language Resources and Technologies for Slovene'' (Grant P6-0411), and the Research Infrastructure DARIAH-SI (I0-E007), all funded by the ARIS Slovenian Research and Innovation Agency.

The authors acknowledge the OSCARS project, and its ParlaCAP cascading grant project, which has received funding from the European Commission’s Horizon Europe Research and Innovation programme under grant agreement No. 101129751.




\bibliographystyle{lrec2026-natbib}
\bibliography{lrec2026-example}


\section{Appendix}
\label{sec:appendix}

In the following sections, we provide the descriptions of the CAP labels that were used in manual annotation and automatic annotation with an LLM (Section \ref{subsec:appendix-labels}), the annotation guidelines that were provided to the annotators of the test data (Section \ref{subsec:appendix-guidelines}), and the prompt used to guide the large language model in the automatic annotation of training data (Section \ref{subsec:appendix-prompt}).

\subsection{CAP Labels Description}
\label{subsec:appendix-labels}

Tables \ref{tab:cap_topics}, \ref{tab:cap_topics2} and \ref{tab:cap_topics3} provide the descriptions of the major CAP topics. These descriptions were used for manual annotation and for automatic annotation of training data with a large language model. They were developed based on the descriptions of CAP subtopics in the Master Codebook~\citep{bevan2019gone}\footnote{\url{https://www.comparativeagendas.net/pages/master-codebook}}, and were further improved based on the Croatian CAP guidelines \citep{sirinic_2019} and expert input.

\begin{table*}[!ht]
\begin{center}
\begin{tabularx}{\textwidth}{|p{0.2\textwidth}|X|}
      \hline
\textbf{Major Topic} & \textbf{Description} \\
\hline
Macroeconomics (1) & Issues related to domestic macroeconomic policy, such as the state and prospect of the national economy, economic policy, inflation, interest rates, monetary policy, cost of living, unemployment rate, national budget, public debt, price control, tax enforcement, industrial revitalization and growth. \\
\hline
Civil Rights (2) & Issues related to civil rights and minority rights, discrimination towards races, gender, sexual orientation, handicap, and other minorities, voting rights, freedom of speech, religious freedoms, privacy rights, protection of personal data, abortion rights, anti-government activity groups (e.g., local insurgency groups), religion and the Church. \\
\hline
Health (3) & Issues related to health care, health care reforms, health insurance, drug industry, medical facilities, medical workers, disease prevention, treatment, and health promotion, drug and alcohol abuse, mental health, research in medicine, medical liability and unfair medical practices. \\
\hline
Agriculture (4) & Issues related to agriculture policy, fishing, agricultural foreign trade, food marketing, subsidies to farmers, food inspection and safety, animal and crop disease, pest control and pesticide regulation, welfare for animals in farms, pets, veterinary medicine, agricultural research. \\
\hline
Labor (5) & Issues related to labor, employment, employment programs, employee benefits, pensions and retirement accounts, minimum wage, labor law, job training, labor unions, worker safety and protection, youth employment and seasonal workers. \\
\hline
Education (6) & Issues related to educational policies, primary and secondary schools, student loans and education finance, the regulation of colleges and universities, school reforms, teachers, vocational training, evening schools, safety in schools, efforts to improve educational standards, and issues related to libraries, dictionaries, teaching material, research in education. \\
\hline
Environment (7) & Issues related to environmental policy, drinking water safety, all kinds of pollution (air, noise, soil), waste disposal, recycling, climate change, outdoor environmental hazards (e.g., asbestos), species and forest protection, marine and freshwater environment, hunting, regulation of laboratory or performance animals, land and water resource conservation, research in environmental technology. \\
\hline
Energy (8) & Issues related to energy policy, electricity, regulation of electrical utilities, nuclear energy and disposal of nuclear waste, natural gas and oil, drilling, oil spills, oil and gas prices, heat supply, shortages and gasoline regulation, coal production, alternative and renewable energy, energy conservation and energy efficiency, energy research. \\
\hline
Immigration (9) & Issues related to immigration, refugees, and citizenship, integration issues, regulation of residence permits, asylum applications; criminal offences and diseases caused by immigration. \\
\hline
Transportation (10) & Issues related to mass transportation construction and regulation, bus transport, regulation related to motor vehicles, road construction, maintenance and safety, parking facilities, traffic accidents statistics, air travel, rail travel, rail freight, maritime transportation, inland waterways and channels, transportation research and development. \\
\hline
\end{tabularx}
\caption{Descriptions of major CAP Topics used in this study.}
\label{tab:cap_topics}
 \end{center}
\end{table*}

\begin{table*}[!ht]
\begin{center}
\begin{tabularx}{\textwidth}{|p{0.2\textwidth}|X|}
\hline
\textbf{Major Topic} & \textbf{Description} \\
\hline
Law and Crime (12) & Issues related to the control, prevention, and impact of crime; all law enforcement agencies, including border and customs, police, court system, prison system; terrorism, white collar crime, counterfeiting and fraud, cyber-crime, drug trafficking, domestic violence, child welfare, family law, juvenile crime. \\
\hline
Social Welfare (13) & Issues related to social welfare policy, the Ministry of Social Affairs, social services, poverty assistance for low-income families and for the elderly, parental leave and child care, assistance for people with physical or mental disabilities, including early retirement pension, discounts on public services, volunteer associations (e.g., Red Cross), charities, and youth organizations. \\
\hline
Housing (14) & Issues related to housing, urban affairs and community development, housing market, property tax, spatial planning, rural development, location permits, construction inspection, illegal construction, industrial and commercial building issues, national housing policy, housing for low-income individuals, rental housing, housing for the elderly, e.g., nursing homes, housing for the homeless and efforts to reduce homelessness, research related to housing. \\
\hline
Domestic Commerce (15) & Issues related to banking, finance and internal commerce, including stock exchange, investments, consumer finance, mortgages, credit cards, insurance availability and cost, accounting regulation, personal, commercial, and municipal bankruptcies, programs to promote small businesses, copyrights and patents, intellectual property, natural disaster preparedness and relief, consumer safety; regulation and promotion of tourism, sports, gambling, and personal fitness; domestic commerce research. \\
\hline
Defense (16) & Issues related to defense policy, military intelligence, espionage, weapons, military personnel, reserve forces, military buildings, military courts, nuclear weapons, civil defense, including firefighters and mountain rescue services, homeland security, military aid or arms sales to other countries, prisoners of war and collateral damage to civilian populations, military nuclear and hazardous waste disposal and military environmental compliance, defense alliances and agreements, direct foreign military operations, claims against military, defense research. \\
\hline
Technology (17) & Issues related to science and technology transfer and international science cooperation, research policy, government space programs and space exploration, telephones and telecommunication regulation, broadcast media (television, radio, newspapers, films), weather forecasting, geological surveys, computer industry, cyber security. \\
\hline
Foreign Trade (18) & Issues related to foreign trade, trade negotiations, free trade agreements, import regulation, export promotion and regulation, subsidies, private business investment and corporate development, competitiveness, exchange rates, the strength of national currency in comparison to other currencies, foreign investment and sales of companies abroad. \\
\hline
International Affairs (19) & Issues related to international affairs, foreign policy and relations to other countries, issues related to the Ministry of Foreign Affairs, foreign aid, international agreements (such as Kyoto agreement on the environment, the Schengen agreement), international organizations (including United Nations, UNESCO, International Olympic Committee, International Criminal Court), NGOs, issues related to diplomacy, embassies, citizens abroad; issues related to border control; issues related to international finance, including the World Bank and International Monetary Fund, the financial situation of the EU; issues related to a foreign country that do not impact the home country; issues related to human rights in other countries, international terrorism. \\
\hline
\end{tabularx}
\caption{Second part of the descriptions of major CAP Topics used in this study.}
\label{tab:cap_topics2}
 \end{center}
\end{table*}

\begin{table*}[!ht]
\begin{center}
\begin{tabularx}{\textwidth}{|p{0.2\textwidth}|X|}
\hline
\textbf{Major Topic} & \textbf{Description} \\
\hline
Government Operations (20) & Issues related to general government operations, the work of multiple departments, public employees, postal services, nominations and appointments, national mints, medals, and commemorative coins, management of government property, government procurement and contractors, public scandal and impeachment, claims against the government, the state inspectorate and audit, anti-corruption policies, regulation of political campaigns, political advertising and voter registration, census and statistics collection by government; issues related to local government, capital city and municipalities, including decentralization; issues related to national holidays. \\
\hline
Public Lands (21) & Issues related to national parks, memorials, historic sites, and protected areas, including the management and staffing of cultural sites; museums; use of public lands and forests, establishment and management of harbors and marinas; issues related to flood control, forest fires, livestock grazing. \\
\hline
Culture (23) & Issues related to cultural policies, Ministry of Culture, public spending on culture, cultural employees, issues related to support of theatres and artists; allocation of funds from the national lottery, issues related to cultural heritage. \\
\hline
Other (0) & Other topics not mentioning policy agendas, including the procedures of parliamentary meetings, e.g., points of order, voting procedures, meeting logistics; interpersonal speech, e.g., greetings, personal stories, tributes, interjections, arguments between the members; rhetorical speech, e.g., jokes, literary references. \\
\hline
\end{tabularx}
\caption{Third part of the descriptions of major CAP Topics used in this study.}
\label{tab:cap_topics3}
 \end{center}
\end{table*}

\clearpage
\subsection{Annotation Guidelines}
\label{subsec:appendix-guidelines}

The following guidelines were provided to the annotators of English, Croatian, Serbian and Bosnian test datasets, along with the label descriptions (see Section \ref{subsec:appendix-labels}).

\textit{Your task is to classify the provided text into a policy agenda topic label, meaning that you need to recognize what is the predominant topic of the text. The labels and their descriptions are provided above.}

\textit{You are provided with excerpts from parliamentary speeches from the Croatian, Serbian, Bosnian or British parliament in Croatian, Serbian, Bosnian or English language.}

\textit{Always provide a label, even if you are not sure.}

\textit{Follow the following rule: if the speech mentions a policy area and a policy instrument (e.g., taxes, laws), pick the label based on the area, not the instrument (e.g., annotate mortgage tax changes with 14 (Housing), law on education with 6 (Education)).}

\textit{Additional label for impossibly hard cases}

\textit{We are interested only in reasonable and informative instances. That is why we added to the list an additional category for you to "discard" the text:
- "do not know": if it is impossible for you to decide under which label this instance fits.}

\subsection{Prompt}
\label{subsec:appendix-prompt}

\begin{figure*}[ht]
\begin{center}
    \includegraphics[width=\textwidth]{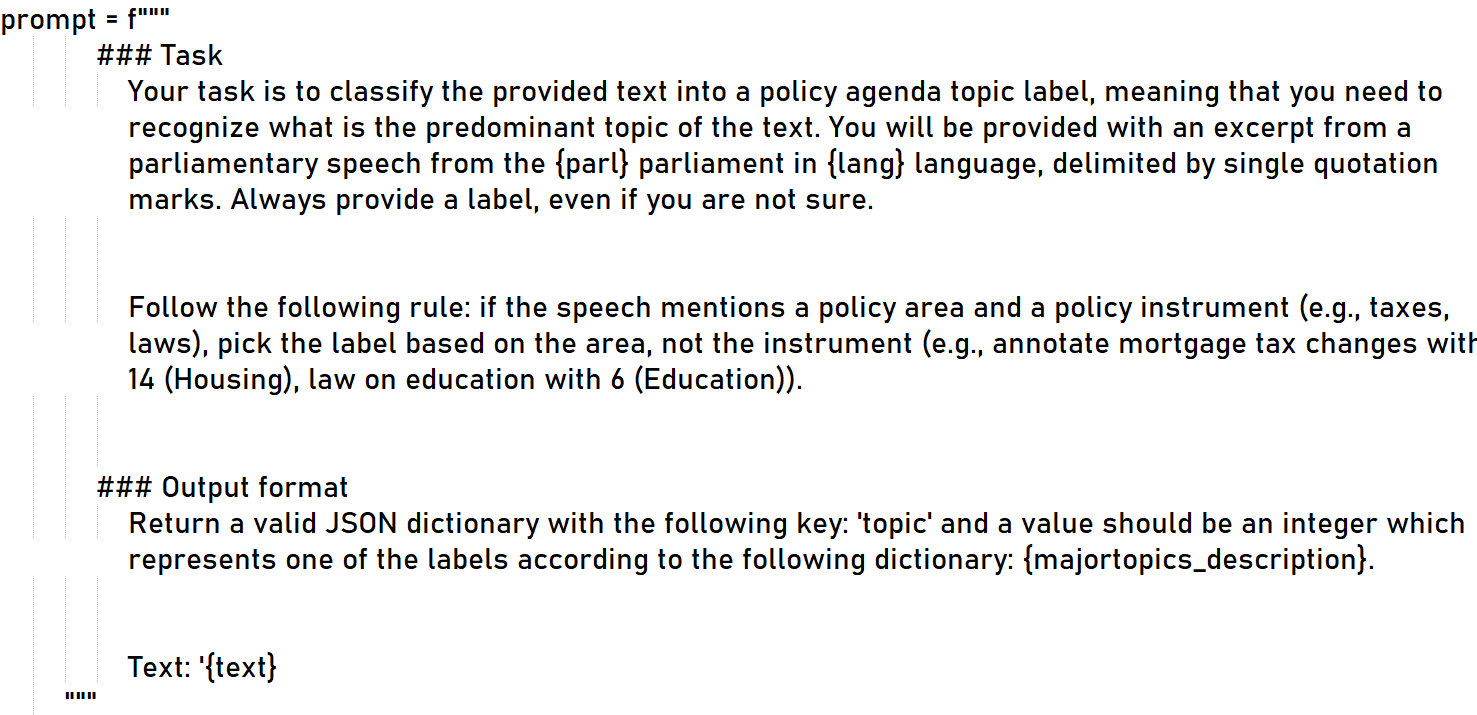}
    \caption{Prompt used for the automatic annotation of parliamentary speeches with an LLM following the CAP schema.}
    \label{fig:prompt}
\end{center}
\end{figure*}

In Figure \ref{fig:prompt}, we present the prompt used to guide the large language model in the automatic annotation of parliamentary speeches with CAP policy topics.

\end{document}